\documentclass[10pt,letterpaper]{article}
\usepackage{apacite}
\usepackage{color,soul}
\usepackage{times}
\usepackage{natbib}
\setcitestyle{authoryear,open={(},close={)}}
\usepackage{pslatex}
\usepackage[table,dvipsnames]{xcolor}
\usepackage{graphicx}
\usepackage{amssymb,amsfonts,amsmath}
\usepackage{subcaption}
\usepackage{mathtools}
\usepackage[normalem]{ulem}
\usepackage{paralist}
\usepackage{tipa}
\usepackage{threeparttable}
\usepackage{booktabs}
\usepackage{multirow}
\usepackage{url}
\usepackage{authblk}
\usepackage{chngpage}

\newcommand\eg{\textit{e.g.,~}}
\newcommand\ie{\textit{i.e.,~}}
\newcommand\MOT{\textsc{mot}}
\newcommand\CHI{\textsc{chi}}

\title{Child-directed Listening:\\How Caregiver Inference Enables Children's Early Verbal Communication}
 
\author[1, 3]{\small Stephan C. Meylan}
\author[2]{\small Ruthe Foushee}
\author[3]{\small Elika Bergelson}
\author[1]{\small Roger P. Levy}

\affil[1]{\footnotesize Department of Brain and Cognitive Sciences, MIT (\{smeylan, rplevy\}@mit.edu)}
\affil[2]{\footnotesize Department of Psychology, University of Chicago (foushee@uchicago.edu)}
\affil[3]{\footnotesize Department of Psychology and Neuroscience, Duke University (elika.bergelson@duke.edu)} 

\begin{document}

\maketitle

\begin{abstract}
How do adults understand children's speech?
Children's productions over the course of language development often bear little resemblance to typical adult pronunciations, yet caregivers nonetheless reliably recover meaning from them. 
Here, we employ a suite of Bayesian models of spoken word recognition to understand how adults overcome the noisiness of child language, showing that communicative success between children and adults relies heavily on adult inferential processes. %labor
By evaluating competing models on phonetically-annotated corpora, we show that adults' recovered meanings are best predicted by prior expectations fitted specifically to the child language environment, rather than to typical adult-adult language. 
After quantifying the contribution of this ``child-directed listening'' over developmental time, we discuss the consequences for theories of language acquisition, as well as the implications for commonly-used methods for assessing children's linguistic proficiency.

\end{abstract}

\section{Introduction}
The past five decades have seen extensive research dedicated to characterizing how adults speak to infants and young children \citep{snow1977talking, soderstrom2007beyond}, and to investigating the degree to which adults’ \textit{child-directed speech} directly supports language learning \citep{golinkoff2015baby}. 
% can remove sentence if need to buy space:
% This literature has primarily focused on how adults' speech to children % + input studies
% \begin{inparaenum}[(a)]
%     \item elicits infants' attention through exaggerated prosody and affect \citep[\eg][]{grafestes2013infant},
%     \item simplifies the learning problem by providing children with calibrated training data \citep[\eg][]{kuhl1997cross,eavesEtAl2016}, 
%     and
%     \item helps children identify their own speech errors and expand on their verbal productions through adult feedback \cite[\eg][]{clark2014pragmatics}. 
% \end{inparaenum} 
By contrast, how caregivers understand the communicative acts of young children --- \textit{child-directed listening} (\textsc{cdl}) --- has received far less attention. 
In this paper, we investigate how English-speaking adults interpret English-learning children's verbal productions, making meaning out of vocalizations that are often perceptually distant from targets in the adult language (\eg /wid/ for \textit{read}; see Table~\ref{example_table}A).

This characterization of adults' role in conversations with young learners dovetails with ``noisy-channel'' accounts of spoken language interpretation, which provide a framework for describing how listeners overcome imperfect acoustic information, verbal ambiguity, distractions, and speaker variability present in everyday conversation \citep{levy2008noisy, shannon1951, gibsonBergenPiantadosi2013}. To recover meanings from highly noisy input, adult listeners rely on their expectations about what speakers are likely to say, combined with the perceptual similarity between what the listener heard and guesses as to what the speaker might intend. 
We argue that child language represents a ``noisier-than-usual'' channel, where adults must use expectations fitted to the child language environment to recover meaning from child productions. %, rather than to adult-adult speech, 
That is, while hearing /wid/ might typically suggest \textit{weed} or \textit{wheat} as a speaker's intended word (based solely on acoustic information), an adult caregiver might instead recover \textit{read} as the intended word from a child speaker.\footnote{One intriguing deviation from the classic noisy-channel setup is that adults may ``recover'' messages when children do not intend to communicate anything at all (\ie drawing from a noise distribution).}

In what follows, we seek evidence for the role of child-directed listening in language development. 
We present a computational framework to predict what adults are likely to recover from children's imperfect speech, and compare it to what adults \textit{actually} recovered.
As a proxy for caregivers' realtime interpretations, we use the orthographic annotations made by trained in-lab transcribers of spontaneous at-home child language recordings. % don't know kind of model
This approach allows us to characterize the utility of adult listeners' expectations, versus the acoustic/phonetic signal produced by the child. 
To capture the degree to which listening is truly \textit{child}-directed (\ie distinct from adult-directed listening), we compare the utility of expectations tuned on large-scale adult corpora, versus expectations tailored to reflect the child language environment. %By quantifying these factors, our analyses speak to a potential developmental process on the part of adults in the course of language development --- that is, how adults learn to listen. 

% In line with this perspective on caregiver-child interactions, we predict: 
% \begin{enumerate}
% \item A decrease in communicative failures over time
% \item Children produce more adult-like articulations over time 
% \item Context + Child Speaker model will best predict the recoveries made by caregivers 
% \item Worse recognition performance for any of the other models (all used with the same likelihood)
% \item Decreasing contribution of caregiver priors as child articulation improves
% \end{enumerate}
\section{Task and Modeling Setup}
% comm success and failure not clear
%Task: Predict the word corresponding to a word (or monosyllabic segment) for child productions in CHILDES. Include both communicative successes (where the word has a gloss) and failures (where the word is transcribed as yyy).
We focus here on the adult listener's task of recovering meaning from noisy child productions.
Specifically, we look at a large set of phonetically-transcribed productions (\eg /\textipa{A}\textipa{@} w\textipa{A}n d\textipa{@} wid/ in Table~\ref{example_table}A) from the Providence corpus \citep{demuthEtAl2006}, and treat the challenge of inferring a word identity in context (here, an orthographic word like \textit{read}) as a \textit{masked word prediction} task (\citealp{devlin2018}).
To combine the contributions of caregiver expectations given the linguistic context with the specific sequence of phonemes produced by the child, we employ a Bayesian model of spoken word recognition in the vein of \citet{norrisMcQueen2008}, which assigns a probability to a candidate word identity $w$ given corresponding perceptual input $d$ in context $c$: 

\vspace{-5mm}
\begin{align}
\label{eq:bayesian_swr} 
P(w|~d,c)  = \frac{P(d|w,c) P(w|c)} {\sum_{w'\in V}{P(d|w',c) P(w'|c)}}
%p(w|d,c)  = \frac{p(d|w) p(w|c)} {\sum_{w'\in \mathbb{W}}{p(d|w') p(w'|c)}}
\end{align}
\vspace{-3mm}
%described in the Introduction, 

This cashes out the intuition that the probability assigned to a candidate word $w$ in spoken word recognition reflects the combination of 
\begin{inparaenum}[(a)]
    \item fit to perceptual data and
    \item linguistic expectations.
\end{inparaenum} 
Fit to perceptual data is evaluated via a likelihood function, $P(d|w,c)$, which reflects the probability that the word $w$ would generate the observed data $d$ in context $c$.
Linguistic expectations are captured in the prior, $P(w|c)$, or the anticipated probability of the word in context $c$, absent any perceptual data.
The denominator in Equation~\ref{eq:bayesian_swr} reflects the summed strength of \textit{all} competitor words $w'$ in the candidate vocabulary $V$.
Thus, the predictions derived from the model (a \textit{posterior}) constitute a probability distribution over candidate words, with highly favored interpretations receiving more of the probability mass than disfavored ones.
%eb VERY clear!!

%MANIPULATING THE PRIOR WHY HOW
Our principal goal is to find a model that best simulates how adults understand children. We discuss the likelihood and prior of the set of models under consideration in turn. 
All models used the same likelihood, derived from measures of pairwise string similarity a phonemic transcription of the child's production and phonemic forms of all candidate words (translated into the International Phonetic Alphabet, IPA, via a dictionary of conventional English pronunciations). %\eg /\textipa{\*r}id/ for \textit{read}). - RF cut for space
To illustrate, given the transcribed production /wid/, the likelihood term for the candidate word \textit{weed} (citation phonetic form /wid/) will be higher than the likelihood term for the candidate word \textit{read} (where the citation phonetic form /\textipa{\*r}id/ differs by one phoneme). %RF - still true with explicit consideration of context?

However, the inferential process sketched in Equation~\ref{eq:bayesian_swr_likelihood} foreshadows the inadequacy of the acoustic signal alone: if children often produce noisy, idiosyncratic phoneme sequences, the prior must do more ``work.'' 
The priors we evaluate take the form of probabilistic language models: computational models that return a probability distribution over word guesses, based on the surrounding linguistic context (Table \ref{example_table}C). 
When priors from each model are combined with the likelihood, they yield posterior distributions (Table \ref{example_table}D).
%Models tested here include ``off-the-shelf'' and CHILDES-fine-tuned variants of BERT \citep{devlin2018}, a recent transformer-based neural architecture, as well as a simple smoothed unigram model based on CHILDES word frequencies. We also evaluate a naive flat prior which assigns all words the same prior probability, as might be expected if adults are only using the phonetic signal to recover meanings from children's utterances.

%eb: not clear to naive reader what BERT is or what you mean by fine-tuned variants;  you  spell this all out later i see, so i'd zoom out up here and not give model specifics that'll be confusing before you've defined them
%sm +rf: fixed -- punted on BERT specifics til Methods

Here, we take advantage of a distinction within the transcripts of caregiver-child speech in the PhonBank database \citep{rose2014phonBank}, which allows us to evaluate competing models on two different dimensions. %rj introduce PhonBank
First, we evaluate models in their ability to reproduce the specific words recovered by annotators.\footnote{We cannot know whether the word recovered by an annotator was the word intended by the child speaker.} 
This analysis focuses specifically on what we term \textit{communicative successes} (Table \ref{example_table}A) --- instances where a phoneme sequence was not only phonemically transcribed (PhonBank \texttt{\%phon} tier), but also received a \textit{gloss}, or orthographic transcription.
This allows us to assess the probability that each model assigns to the annotator-recovered word, with the best model being the one that assigns the highest average probability (alternatively, the lowest \textit{surprisal}, or negative log probability\footnote{For statistics-oriented readers, this is the per-instance log-likelihood of the data under the model.}) to the glosses.

Second, we test whether models can predict when a child's production will \textit{not} receive a gloss (reflecting the annotator's uncertainty as to the child's intended word). 
This analysis relies on the communicative successes described above, as well as so-called \textit{communicative failures} --- instances where phoneme sequences are transcribed, but lack a gloss, due to difficulty in identifying the child's intended word (Table \ref{example_table}B).
In the absence of an annotator-recovered word, surprisal cannot be calculated.
Instead, we measure the ``peakedness'' of the guesses regarding word identity by calculating the  \textit{information entropy} of the posterior distribution,

\vspace{-5mm}
\begin{align}
\label{eq:entropy_def} 
H(X) = -\sum_{i=1}^n{P(x_i)\log P(x_i)},
\end{align}
\vspace{-3mm}

\noindent where $P(x_i)$ is the probability of the $i$th candidate word. 
This provides us with a concise index of uncertainty: if posterior probability mass is centered on one or a few guesses for a given phoneme sequence, then entropy will be low; if the posterior is split across many candidate guesses, then entropy will be high. 
The best model under this analysis will be the one most able to discriminate failures from successes on the basis of entropy.
We measure this with the \textit{receiver operating characteristic}, or ROC, which measures the diagnostic ability of a classifier over the range of possible thresholds. %cite?

% measure the relative contribution of the likelihood and prior in meeting goals 1 and 2
In the third analysis, we quantify how much the estimate of word identity changes as a function of 1) conditioning on context (using a fitted prior) 2) conditioning on data (the posterior when using a uniform prior), or 3) conditioning on \textit{both} context \textit{and} data (the posteriors reflecting the fitted priors).
As a baseline for comparison, we start with a uniform prior, where all words in the vocabulary are equiprobable.
We then measure the per-word average \textit{information gain}, or Kullback-Liebler divergence, between that uniform prior distribution and each of the distributions identified above.
Information gain can be interpreted as a measure of \textit{entropy reduction}, corresponding to the difference between the uniform prior and the somewhat more peaked estimates of word identity under the fitted priors, and the (usually) yet more peaked estimates under the posteriors.
If the models are using the perceptual signal to identify words, then the prior information gain will be small in comparison to the posterior information gain.
If, by contrast, caregivers are relying heavily on their prior expectations, then the prior information gain will be larger with respect to the posterior information gain.

A further question is how these measures of information gain will track with developmental time. 
We expect prior information gain to \textit{increase} over developmental time: as the child says more words in the surrounding context, the priors can better constrain guesses for the masked words (placing more mass on a smaller set of words, reflected in lower entropy).
At the same time, as children's productions approximate conventional pronunciations, we expect to see an increase in posterior information gain.
It remains to be seen how these two quantities will interact.
%Likewise the information gain for the posterior should increase even more over developmental time than the contribution of the prior, reflecting an increase in the likelihood of observed productions as children's articulation improves.
%Depending on the relative strength of these two effects...
%eb this last para seems like an afterthought but actually seems to contain a critical point: clarify.

%eb:  a lot  to  unpack in this preceding paragraph for naive reader. i know you mean that all kid prod's got phon tier but sometimes no gloss tier i.e. coder wasn't sure what they meant to say( right?)  but not sure how your reader would know that; could keep it higher level here 
%sm + rf: kept it higher level for most of i, but provide additional details for the analysis

%eb: give an example to make this concrete?
% sm + rf: added refs to table
%also what's a segment?
% sm + rf: switched to phoneme sequence

\begin{table}[]
\centering
\begin{adjustwidth}{-8em}{-8em}
\begin{threeparttable}
\caption{Examples of communicative success and failure, with samples from highest-ranked prior and posterior candidates. \label{example_table}}
\begin{tabular}{lllcll}
\toprule
\multirow{7}{*}{\rotatebox{90}{PhonBank transcript~~~~~}} & \multicolumn{2}{c}{\textbf{A. Communicative Success}\tnote{$^\dagger$}} 
    &~& \multicolumn{2}{c}{\textbf{B. Communicative Failure}\tnote{$^\dagger$}}\\[.25ex]            
\cline{2-3} \cline{5-6}\\[-1.5ex] 

&\textcolor{gray}{\MOT} & \textcolor{gray}{this is}                                             
    &~& 
        \textcolor{gray}{\MOT} & \textcolor{gray}{do you want ta put some beans in your eggs?}             \\
& \textcolor{gray}{\MOT} & \textcolor{gray}{you want mamma let's see}                                         
    &~& 
        \textcolor{gray}{\CHI} & \textcolor{gray}{no} \\[.5ex]
\multicolumn{1}{r}{\texttt{\%phon}} & \CHI & / {\textipa{A}\textipa{@}~~w\textipa{A}n~~d\textipa{@}~~\fbox{wid\tnote{$^*$}}} /  &~& 
    \CHI & / ju~~~~m\textipa{E}\textipa{I}k~~yo\textipa{@}\textipa{\*r}~~~\fbox{f\textipa{E}t\tnote{$^*$}} /\\
\multicolumn{1}{r}{\textit{gloss}}  && \textit{~~I~~~~want~~to~~\textless{read}\textgreater}
    &~& 
         & \textit{~you~~make~~your}~$<$\textit{unintelligible}$>$ $\rightarrow$ \texttt{yyy}      \\[.5ex]
& \textcolor{gray}{\MOT} & \textcolor{gray}{okay that's fine}                                                   
    &~& 
        \textcolor{gray}{\MOT} & \textcolor{gray}{can I make one?}                                      \\
& \textcolor{gray}{\MOT} & \textcolor{gray}{okay mommy's gonna pick out a book}                     
    &~& 
        \textcolor{gray}{\MOT} & \textcolor{gray}{no} \\[.5ex]
\midrule\\[-2ex]
%\hline
\textbf{Language Model} & \multicolumn{5}{c}{\textbf{C. Best \textsc{Prior} Guesses for} \fbox{wid / f\textipa{E}t}} \\[.5ex] 
\cline{2-6}\\[-2ex]
\textsc{cdl+context}\tnote{$^\ddagger$}    & 
    \multicolumn{2}{l}{see (.86)~~~look (.03)~~~go (.02)~play (.01)}                                    
    &~& 
        \multicolumn{2}{l}{own (.74)~~~~house (.01)~~~shapes (.01)~~~~friends (.01)}            \\
\textsc{bert+context}\tnote{$^\ddagger$}      & 
    \multicolumn{2}{l}{\textit{read} (.49) see (.28)~~play (.04) know (.04)}                       
    &~& 
        \multicolumn{2}{l}{own (.25) choice (.24) point (.04) bed (.03) call (.03)}                \\
\textsc{childes-1gram}            & 
    \multicolumn{2}{l}{I~~~~(.04)~~~~a (.03)~~~~~the (.03)~~~~yeah (.03)} 
    &~& 
        \multicolumn{2}{l}{I (.04)~~~~~~a (.03)~~~~~the (.03)~~~~~yeah (.03)~~~~~~it (.02)}
        \\[.5ex] 
\midrule
& \multicolumn{5}{c}{\textbf{D. Best \textsc{Posterior} Guesses for} \fbox{wid / f\textipa{E}t}} \\[.5ex]
\cline{2-6}\\[-2ex]
\textsc{cdl+context}\tnote{$^\ddagger$}    & 
    \multicolumn{2}{l}{see (.967) watch (.012) \textit{read} (.005) look (.001)}                                
    &~& 
        \multicolumn{2}{l}{own (.59)~ feet (.27)~ foot (.02)~food (.01)~~hat (0.01)}            \\
\textsc{bert+context}\tnote{$^\ddagger$}      & 
    \multicolumn{2}{l}{\textit{read} (.61)~~~~see (.35)~~~~watch (.01)~~~~hear (.01)}                                     
    &~& 
        \multicolumn{2}{l}{bet (.31)~~~own (.24)~~cut (.06)~~~shot (.04)~~~bed (.03)}         \\
\textsc{childes-1gram}            & 
    \multicolumn{2}{l}{we (.34) ~~~need (.11) ~~~and (.06) ~~~would (.04)} 
        &~& 
            \multicolumn{2}{l}{it (.15)~~~~that (.11)~~~fit (.06)~~~ what (.06)~~~~feet (.05)}       \\[.5ex] 
\bottomrule\\[-2ex]
\end{tabular}
\begin{tablenotes}
    \item[$^*$] masked phoneme sequence \hspace{10pt} 
    $^\dagger$ \MOT = Mother, \CHI = Child \hspace{10pt} 
    $^\ddagger$ Model considers $+/-20$ utterances of surrounding context.
    %\item[$^\ddagger$] Language model includes $+/-20$ utterances of surrounding linguistic context.
\end{tablenotes}
\end{threeparttable}
\end{adjustwidth}
\end{table}

%eb: i like this table, esp that adult bert guesses ' i want to die'. but what's the strikethru mean? also could be confusing that  # models != # sets of priors for a reader who's not quite following

% justified guesses?

%\begin{table}[h]
%    \centering
%    \begin{tabular}{l|cccc|cccc|}
%    & \multicolumn{4}{c}{Communicative Success} & \multicolumn{4}{c}{Model Predictions}\\
%    \hline
%        gloss & \textit{I} & \textit{want} & \textit{to} & \sout{\textit{read}}  & CHI-& & & \\
%        phon. & \textipa{A}\textipa{@} & w\textipa{A}n & d\textipa{@} & wid & BERT-2 & & & \\
%        model & \textipa{A}\textipa{@} & w\textipa{A}nt & tu & \fbox{\textipa{r}id} & BERT-9 & & & \\
%        \crule{}
%    \end{tabular}
%    \caption{Caption}
%    \label{tab:my_label}
%\end{table}
%1032328	I want to read	ɑə	ɑə	16928243
%1032329	I want to read	wɑn	wɑnt	16928243
%1032330	I want to read	də	tu	16928243
%1032331	I want to read	wid	ɹid	16928243

\section{Methods and Model Details}

We test several language models in their ability to predict adult caregivers' interpretations of children's linguistic and proto-linguistic vocalizations in the Providence corpus \citep{demuthEtAl2006}.
Utterances and phonological transcripts with both phonemic and orthographic transcription were retrieved through childes-db 2020.1. \citep{sanchez2019}. %Phonbank format

\subsection{Selecting Communicative Successes and Failures}
We selected as communicative successes all tokens produced by children in the intersection of four criteria:  
\begin{inparaenum}[(1)]
    \item possessing monosyllabic IPA forms (motivated below)
    \item possessing \textit{no} unintelligible (CHILDES code \texttt{xxx}) or phonology-only  (\texttt{yyy}) tokens in the same utterance
    \item whose gloss is extant as a token in BERT (motivated below)  % eb:i have no idea what that noun mash-up means
    \item whose gloss is included in the Carnegie Mellon Pronunciation Dictionary (henceforth CMU dictionary).
\end{inparaenum}
%eb:  you've been using com.success/failure already: do a pass for defining jargon when you first meet it, OR maybe better?, describe it in plain language before you'rr ready to operationalize here
%eb:  the 'we selected' sentence is hard to parse; not clear why 1 syll, not clear what xxx or yyy means or why  its excluded, lexambiguity about the PoS of 'present' led me to parsing difficulties.  maybe better to explain each as you go 
Communicative failures had to meet the first criterion, but must have received the gloss of \texttt{yyy} (with no other \texttt{yyy} or \texttt{xxx} in the same utterance).
Under these definitions, an utterance could contain several communicative successes, but at most one failure. 

\subsection{Candidate Vocabulary} The inventory of candidate words considered by each model was the intersection of 
\begin{inparaenum}[(1)]
    \item{words in the CMU dictionary with one or two syllables and}
    \item{tokens present in BERT  (motivated below)} %eb see previous note+ should explain things on first mention
    \item{tokens that appeared 3 or more times in CHILDES (to limit to words that might reasonably be said in this context).}
\end{inparaenum}
This means that while only one-syllable phoneme sequences were analyzed, two-syllable words were also considered as possible candidate interpretations.
The final inventory of candidates, $V$, included 7,904 words.
We reconcile IPA formats following a procedure detailed in our code.
%eb: again some redundancy with  preceding text and terminology. smooth, simplify, justify
%eb: still true in this second pass, also not clear ~why~ 1 vs 2 shift in this vs. preceding para

\subsection{Priors: Language Models} For each communicative success and failure, we retrieve prior probabilities over candidate words using a suite of probabilistic language models.
As a ``best'' prior architecture, we use BERT \citep{devlin2018}, which has demonstrated extremely competitive performance for single-word completion tasks, including spoken word recognition \citep{salazarEtAl2020}.
By virtue of its attentional mechanisms, BERT is able to effectively model long distance dependencies \citep{jawahar2019}, and capture speech register and discourse-level information.
We compute the probabilities for the masked word $P(w)$ from BERT, using a language modeling head with the \texttt{transformers} library \citep{wolfEtAl2020}.
For each masked phoneme sequence, we take the real-valued vector of predictions corresponding to the model's vocabulary, extract the activations corresponding to the candidate words, and compute the softmax to yield a vector of probabilities over the candidate words (Table \ref{example_table}).

We test an ``off-the-shelf'' model of BERT trained on large quantities of (principally adult-directed) language scraped from the internet, predicting the word from the immediate utterance only (\textsc{BERT+OneUtt}).
We additionally test the predictions of a BERT model meant to best capture adult expectations about children's utterances.
To do this we ``fine-tune'' the above model on adult and child CHILDES utterance glosses, excluding PhonBank.
In fine tuning, a new model is initialized with an ``off-the-shelf'' model, then the weights in the model are updated to best predict masks inserted into a new training set --- in this case, the lines of 80\% of CHILDES transcripts (20\% were held out for model validation).
This fine-tuned model (\textsc{CDL+OneUtt}) should be expected to be more representative of adult linguistic expectations in understanding child speech than the off-the-shelf model for three reasons.
First, it should assign higher probability to words that are common in speech to and from children.
Second, it should assign higher probability to non-sentence fragments, which are ubiquitous in conversational speech but somewhat less prevalent in adult-directed written language. % used to train bert one utt
Third, it may prove capable of developing an expectation for the dyadic, back-and-forth structure of scenes typically captured in transcripts.

In addition to fine-tuning the model, we manipulate whether prior estimates reflect access to the larger discource context as captured by the transcript before and after a phoneme sequence. 
In that these models are meant to be representative of \textit{caregiver} expectations, these models condition the prediction of the masked token on what the caregiver and child \textit{both} say, both before and after the masked token.
We create priors parallel to those above by feeding the models 20 utterances preceding and following each mask (\textsc{CDL+Context} and \textsc{BERT+Context}).

BERT has its own vocabulary, which imposes limitations on the vocabulary in the analysis.
Standard implementations of BERT split longer words into ``word pieces'', or most common repeated sub-sequences.
In English, this often yields morphological segmentation (\eg \textit{fishing} $\rightarrow$ \texttt{fish}~~\texttt{\#\#ing}), but the process is highly noisy.
For the purposes of predicting a masked word, BERT predicts only one word or word piece. %quotes? 
We limit the vocabulary to word-initial word pieces like \textit{fish}, and exclude continuations like \texttt{\#\#ing} from consideration.
This also motivates the choice to predict monosyllabic phoneme sequences, in that the model does not allow us to predict multiple words (which might be contained in \texttt{yyy}).
%eb: just noting again,  i  am r eally  struggling with parsing non-word piece, that sounds like it's a piece that's not a word. hyphenation inconsistent too

In addition to the BERT models, we also test two simpler priors.
The first is a simple smoothed unigram model estimated from counts in CHILDES.
This model, \textsc{CHILDES 1-gram}, assigns probability to all word types proportional to their counts in the same CHILDES dataset used in the \textsc{CDL} models, above.
To account for unseen data, we add a small pseudocount (.001) smoothing to all counts before computing probabilities.
The second is the \textsc{UniformPrior} model, which assigns equal probability to all words ($1/|V|$, where $|V|$is the number of candidates). 
This provides the comparison case of a maximally uninformative prior.

\subsection{Likelihood} For the likelihood, $P(d|w)$, we use a transformation of string edit distance between the phoneme sequence produced by the child and all candidate words.
Specifically, we use exponentiated negative edit distance \citep{levy2008noisy}:

\vspace{-5mm}
\begin{align}
\label{eq:bayesian_swr_likelihood} 
P(d|w) \propto e ^{-\beta~\times~ \text{dist}(d':w',~d)}
%p(w|d,c)  = \frac{p(d|w) p(w|c)} {\sum_{w'\in \mathbb{W}}{p(d|w') p(w'|c)}}
\end{align}
\vspace{-5mm}

\noindent where $dist$ is the Levenshtein distance (minimal number of deletions, insertions and substitutions) between citation form $d'$ for candidate word $w'$, designated here $(d':w')$, and the observed transcription $(d)$.
For the results presented here, we grid sample $\beta$ values between 1 and 6 by $0.1$ increments, and take the value that assigns the highest posterior probability to a sample of 1000 communicative successes across models ($\beta = 3.2$). 
This treatment of edit distance does not take into account phoneme similarity, \ie that certain phonemes are much more perceptually similar.
We propose another more sophisticated likelihood function that captures this in the Discussion.

All model training and analysis code, as well as the fine-tuned model can be accessed at \url{https://osf.io/v7c3e/?view_only=176bb0f538af424da59007c53eff7e05}.

\section{Results}
\subsection{Predicting Adult Recoveries} A comparison of Bayesian speech recognition models reflecting different priors reveals that the {\textsc{cdl+context} prior} assigns the lowest average surprisal (highest average probability) to the recovered word gloss in the transcript.
As Table~\ref{tab:model_surprisal_comparison} reveals, BERT models making use of context perform better than those that do not.
CHILDES-tuned BERT models outperform the respective off-the-shelf BERT models.
All BERT models outperform the \textsc{CHILDES 1gram} model, and all models with fitted priors assign significantly higher probability to the recovered glosses than the \textsc{UniformPrior} model.
These results mean that the model that is 
\begin{inparaenum}[(1)]
    \item fine-tuned to the child environment and
    \item uses the surrounding utterance context 
\end{inparaenum} 
is best able to predict the recoveries made by adults.
%eb: zoom out and tell us what these results means, practically and non-technically  speaking,  for the question they set out to answer;  i'd do this for every key point/para of results
% sm: done

\begin{table}[b!]
\caption{Average prior surprisal on communicative successes from the Providence corpus (lower is better).
The difference in average probability assigned to the actual gloss is $2^{\text{diff}}$, where $\text{diff}$ is the difference between two model scores.$^*$Paired $t$-tests confirm sig. differences between models, $p<10^{-5}$.}
\label{tab:model_surprisal_comparison}
\centering
\begin{tabular}{l|c}
Model & Avg. Prior Surprisal$^*$ (bits) \\
\hline
\textsc{CDL+Context}   & 3.17              \\
\textsc{BERT+Context} & 4.59              \\
\textsc{CDL+OneUtt}         & 5.28             \\
\textsc{BERT+OneUtt}      & 7.09              \\
\textsc{CHILDES 1gram}                 & 8.80 \\
\textsc{UniformPrior}   &  12.95
\end{tabular}
\end{table}

\begin{figure}
\centering
\includegraphics[width=.7\linewidth]{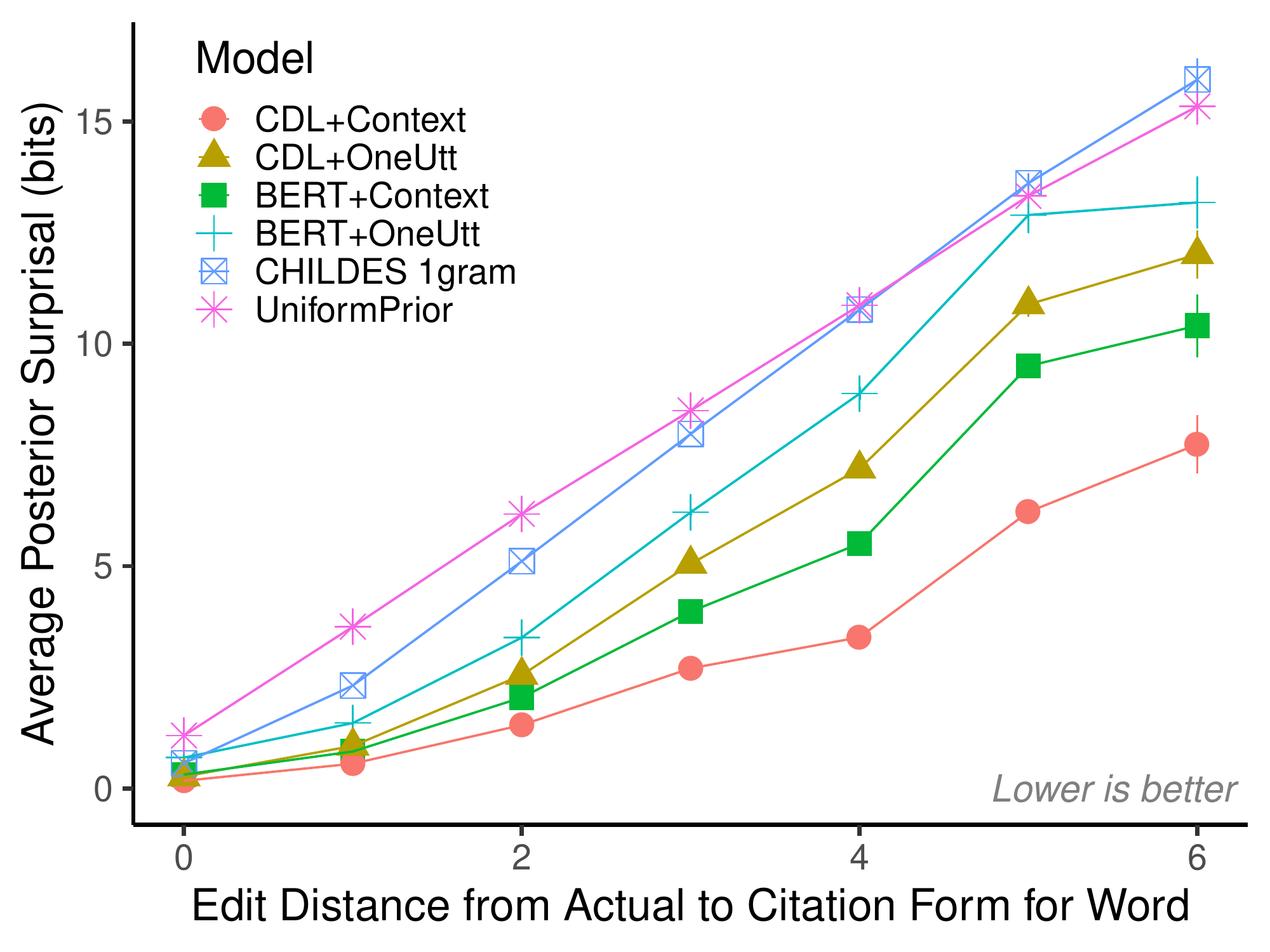}
\caption{
Posterior surprisal (negative log probability) of the recovered meaning for communicative successes.
Error bars indicate standard error of the mean.
}
\vspace{-5mm}
\label{fig:posterior_entropy_by_edit_distance}
\end{figure}

We next investigate how the prior probabilities in the previous analysis combine with likelihoods to predict word identity.
That is, how do the adults' prior expectations support inference when children's productions are more or less adult-like?
Comparing average surprisal across edit distances (Figure~\ref{fig:posterior_entropy_by_edit_distance}) reveals that models using BERT-based priors assign massively higher probability to word identities posited by annotators. 
For child productions that are 2 phonemes away from the citation form ($x = 2$ in Fig. \ref{fig:posterior_entropy_by_edit_distance}), \textsc{CDL+Context} assigns on average a probability of .24 ($2^{-1~\times~surprisal}$) to the correct gloss. 
This compares favorably to .12 under \textsc{BERT+Context}, .08 under \textsc{CDL+OneUtt}, .03 under \textsc{BERT+OneUtt}, .006 under the \textsc{CHILDES 1gram}, and .002 under \textsc{UniformPrior}.
\textsc{CDL+Context} assigns uniformly higher probability (lower surprisal) to the correct word identity, particularly when the phonetic form is more dissimilar (3 or more edits).
This means that priors support recognition more when the perceptual input is noisier.

%eb: ensure probabilities and p-values for significance aren't confusable to reader
% sm fixed

% \begin{figure}
% \includegraphics[width=\linewidth]{figures/ft1_20uttscontext_entropy.pdf}
% \caption{
% Entropy in $p(w)$ for communicative successes vs. failures under a BERT model fine-tuned on CHILDES that uses the surrounding context. 
% }
% \label{fig:ft1_20uttscontext_entropy}
% \end{figure}

\subsection{Predicting communicative failures}
\begin{figure}
\centering
\includegraphics[width=.7\linewidth]{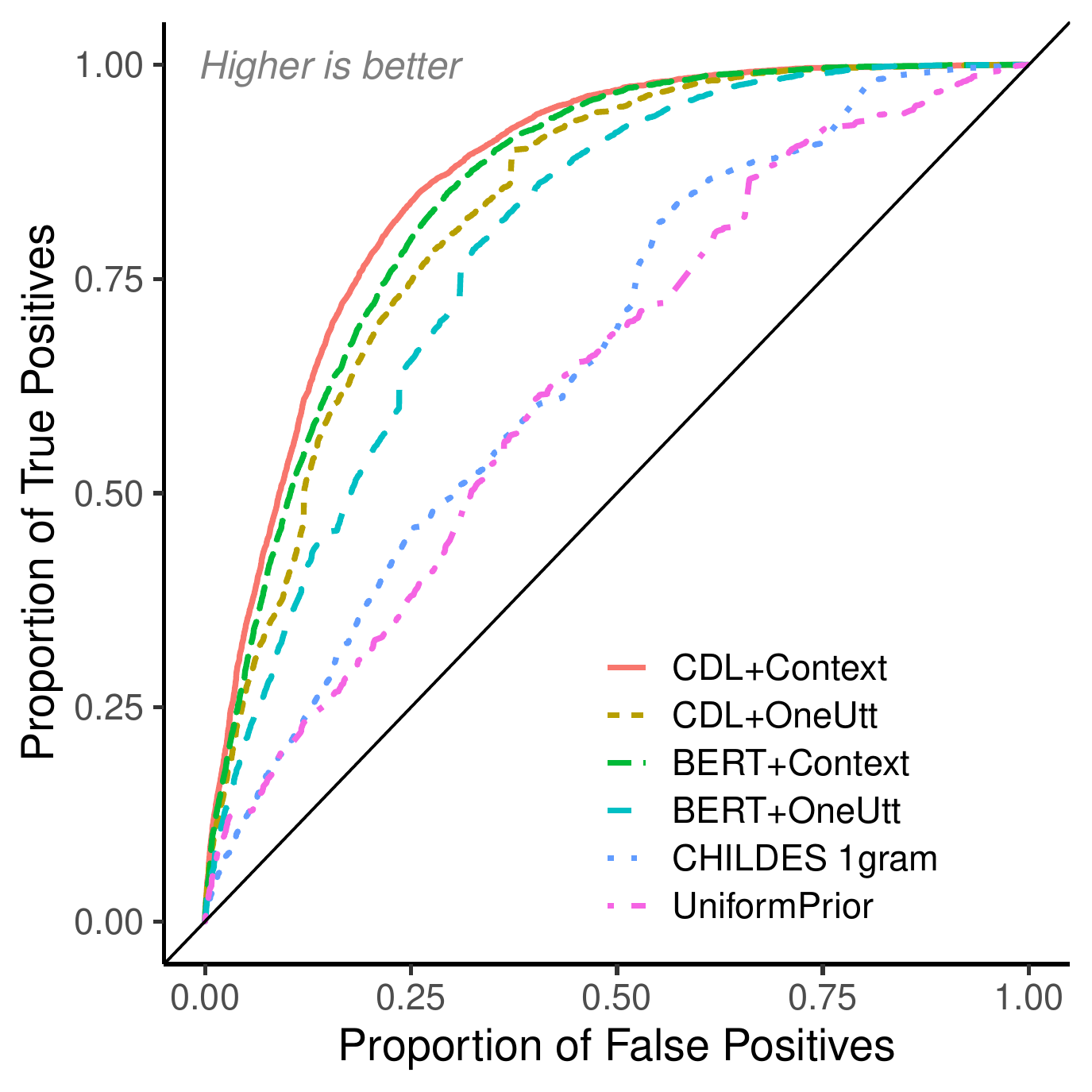}
\vspace{-3mm}
\caption{
Classification performance in predicting communicative failures, as measured by the ROC of posterior entropy. 
The solid line with slope = 1 indicates chance.
The area above this line indicates better classification performance.
}
\vspace{-5mm}
\label{fig:success_by_entropy}
\end{figure}

A separate question is which model best predicts whether a particular phoneme sequence will be a communicative success or failure.
We address this by testing how well posterior entropy under the models can predict communicative failures.   
As with the first analysis, the \textsc{cdl+context} model provides the best trade-off between the prevalence of true positives and false positives (Figure~\ref{fig:success_by_entropy}). 
As both \textsc{UniformPrior} and \textsc{CHILDES 1gram} models assign constant entropy to phoneme sequences (prior probabilities of candidates do not change as a function of context), their posterior entropy \textit{only} reflects the contribution of the perceptual data.
This analysis provides converging evidence that a model that is tuned specifically to child language and uses the surrounding utterance context --- the one that best instantiates child-directed listening --- is best able to replicate adult inferences.

\subsection{Quantifying prior vs. posterior information}
\begin{figure*}
\includegraphics[width=\linewidth]{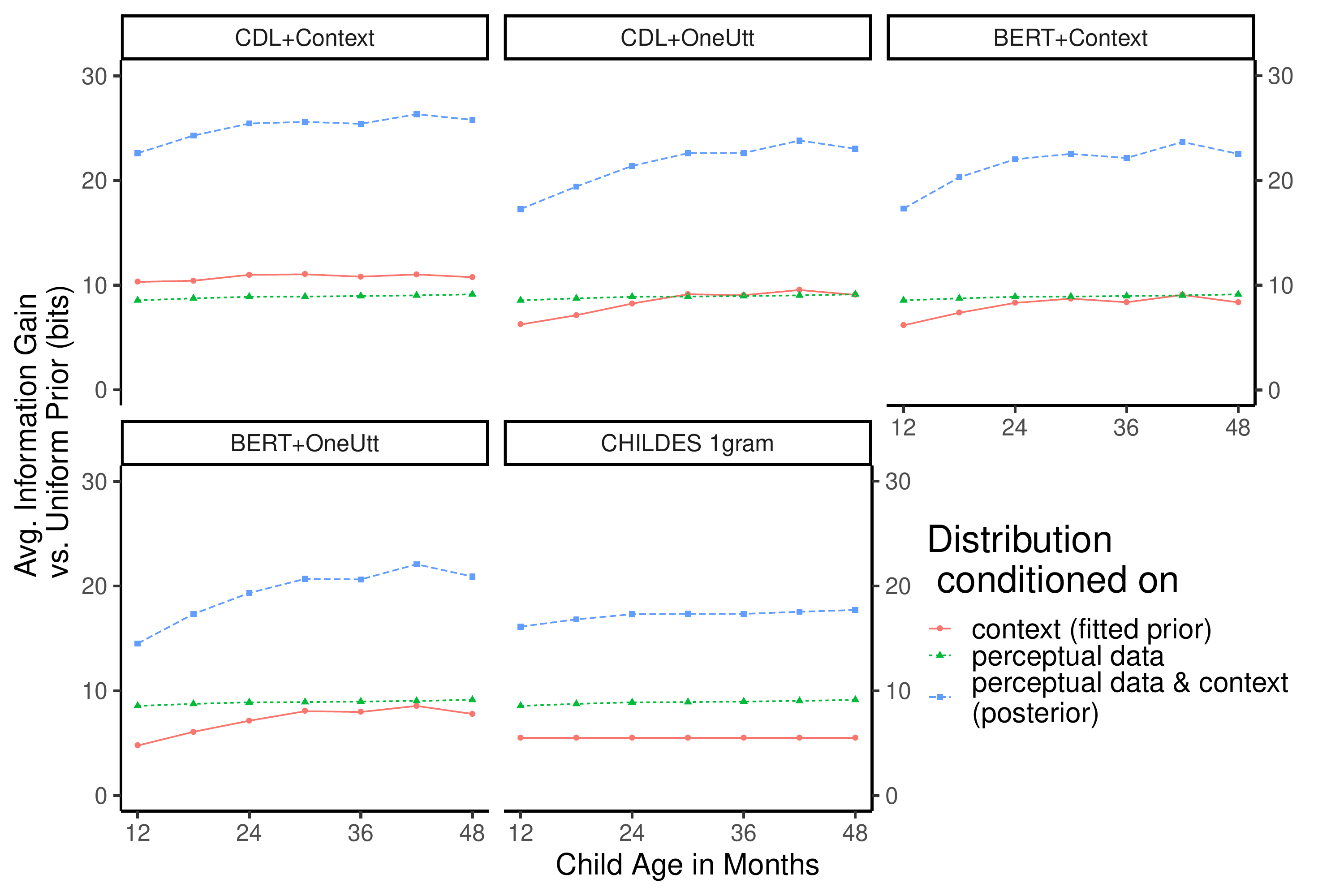}
\vspace{-5mm}
\caption{
Average information gain from conditioning word prediction on context only (red, corresponding to the prior), perceptual data only (green), and context and perceptual data (blue, corresponding to the posterior) relative to a uniform prior.
\label{information_gain}
}
\label{fig:infomration_gain}
\vspace{-5mm}
\end{figure*}

Finally, we quantify the information gain over time in conditioning on context (the fitted priors), conditioning on data (the posterior under the \textsc{UniformPrior} model), and conditioning on both (the posteriors corresponding to the fitted priors).
This analysis shows a larger shift in the probability distribution over candidates (greater information gain) going from the uniform prior to the \textsc{cdl+context} prior compared to going from the uniform prior to its corresponding posterior (red line vs. green line in panel 1 of Figure~\ref{information_gain}).
That is, the prior under the \textsc{CDL+Context} model contributes \textit{more} information (better constrains guesses to word identity) than perceptual information alone.
Contrary to our predictions, we find that the information gain for the prior is relatively constant over time for the CHILDES-fitted models.
This suggests that child-directed listening can helpfully constrain adult listeners' interpretations of children's earliest verbal productions.
As expected, children's improving articulatory abilities result in an increase of all models' posteriors over developmental time, as the likelihood function shared across models is able to contribute more and more to the task of interpretation.
%eb: very helpful reader scaffolding in this last para; might mention this age/artic/interp thing earlier too if space allows

\section{Discussion}
Language development is often characterized in terms of an increasing facility with processes on the side of the learner: developing motor planning, recognizing regularities of linguistic structure at different levels, and relating structure to entities and communicative contexts in the world. 
The current work suggests that early verbal communication depends not only on these well-studied developmental processes, but also on cognitive processes in the minds of adult caregivers.

We note two limitations with the current work before discussing its implications. 
First, the simple measure of edit distance does not capture the perceptual confusability of phonemes: \textit{bug} and \textit{rug} are equally good candidates for \textit{pug}. %standardize italic v quotes
One potential elaboration would be to use a \textit{weighted} edit distance measure that takes into account the perceptual confusability of the phonemes.
For example using a \textit{probabilistic finite state string transducer} would allow assigning edits different ``costs'' according to experimentally-obtained confusion probabilities, \eg \citep{cutlerEtAl2004}. 

Second, we make the simplifying assumption that inferences made by adult annotators in the lab are representative of the inferences made by adult caregivers in the moment, communicating in real time with children.
While the inferential capacities of annotators are likely substantially \textit{less} than those of adult caregivers (who have access to the non-linguistic context, as well as significantly more shared history with the child), research assistants may well be a decent proxy for adult listeners, due to their training as transcribers and exposure to child language.
%eb: that said, RAs with child lang training and who work with lots of kids are likely  decently calibrated relative to some random joe shmo to expect things like lello for yellow , wed for red, or w/e.
% sm: added
Potential differences in the inferential capacities of caregivers relative to other adult ``listeners'' should be tested experimentally.

These results additionally call attention to the interpretation of common methods in child language research.
For example, vocabulary production measures on the Communicative Development Inventories \citep{fenson2007macarthur}, have been historically interpreted as an index of children's vocabulary and articulatory maturity.
However, the current work suggests that successful communication -- adult recognition of a word as a conventional form --- relies additionally on adult inferential processes.
Indeed the measure of a word's ``babiness,'' 
a significant predictor of the order of children's reported vocabulary production, % missing cite?
may reflect the degree to which a word is more likely in child-directed speech compared to adult-directed speech.
%eb: sig pred of what? the babiness var for Mike's CDI stuff was always this magic mojo that needed to be unpacked imho [not necessarily here]
% sm: yes precisely! we should just be able to calculate (p(w|child prior) / p(w|adult prior) and see how it correlates with babiness. Clarified 

%eb:?
%eb: is naive flat prior straw man bc doesn't account even basic things like PoS freq?
%sm: Those expectations ARE the priors. That said, you don't actually get much from pos in child language: utts are short and they are mostly n-v, n-n and v-n sequences. The only real constraint there is that if you have seen a v, you probably won't see another one

% Developmental process on the part of adults \\ 
% ``non-stationarity of the receiver'' \\
%Parental report, eg CDI: interaction of children's abilities and those of adults \\ %in fact, the imprint of the parent/adult ear is everywhere on what we know of kids
%CDI doesn't just reflect child knowledge
%Babiness as a measure of adult reoverability
%Speech recognition for transcription requires special priors

%In so doing, we [bring together developmental and psycholinguistic literatures].
%Zone of proximal development \citep{vygotsky1978mind}: shifting requirements for communication 
%Child is participating in noisy-channel communication; properties of the receiver are critically important

%The question of Feedback in child language \\
%New construal of feedback: successful communication to adults, linked to adult's ability to effect change on the part of the child \\
Furthermore, our data invite a reconstrual of the nature of feedback in early language development.
For example, if we assume that successful communication is itself reinforcing, child-directed \textit{listening} might provide feedback to the child learner even in the absence of child-directed \textit{speech}: a caregiver who interprets a child's production of ``uh'' to mean ``up'' may not \textit{say} anything in response to the child's production, but provides feedback by effecting change on the part of the child when they pick the child up. 
This, in turn, leads to new puzzles: if adult caregivers can help many deficient communicative acts succeed, what presses children to get better?
%eb: YES! well put

Finally, we speculate regarding the role that child-directed listening might contribute to the emergence of language, both on evolutionary timescales and cases of rapid language emergence like Nicaraguan Sign Language.
The current work suggests that successful recovery of meaning from child speech acts reflect not only the inductive biases, linguistic knowledge, and articulatory maturity of speakers, but also the inferential biases of listeners. 
%Pruning
% Cannot speak to how CDL advances language development

% \item <something about predicting CDI: average expectedness in context vs mean articuatory variability in the dataset> Children need to produce more accurate forms for lower probability words in order to be ``given credit''
% vs. null hyptothesis that children will have less accurate forms for lower probability words because they have seen them less often
%role of priors in advancing communicative success, but 

%Broader questions regarding the emergence of language.
%As presented here: chicken and egg.
%Cultural rachet.
%Bootstrapping: NSL
%Listener Expectations as well as speaker knowledge
%Priors to interpret productions in conventionalized ways

\section{Conclusion}
We present a suite of Bayesian models of spoken word recognition to characterize the process of \textit{child-directed listening}, or how adult caregivers find meaning in the noisy and often non-conventional speech productions of young children.
We find that priors capitalizing on recent neural architectures --- when trained specifically on child speech samples, and taking advantage of the greater linguistic context to make predictions --- are best able to simulate adult inferential processes when interpreting noisy child speech. %predict the recoveries made by adult listeners.
This research paves the way for understanding how children learn to employ language as goal-seeking agents in the presence of others. 
% say the  penultimate sentence with more pizzazz and excitement! and i'd get y our hopes out of the last sentence:  just say you ARE paving the way for how children  become mature language users thnx to their interlocutors:)

\vfill\eject
\setcounter{secnumdepth}{0}
\renewcommand*{\bibfont}{\footnotesize}
\bibliography{cdl_cogsci.bib}

\end{document}